\documentclass[lettersize,journal]{IEEEtran}
\usepackage{amsmath,amsfonts}
\usepackage{algorithm}
\usepackage{algpseudocode}
\usepackage{array}
\usepackage[caption=false,font=footnotesize,labelfont=sf,textfont=sf]{subfig}
\usepackage{textcomp}
\usepackage{stfloats}
\usepackage{float}
\usepackage{url}
\usepackage{nomencl}
\usepackage{orcidlink}
\usepackage{verbatim}
\usepackage{epstopdf}
\usepackage{graphicx}
\usepackage{enumerate}
\usepackage{tabularx}
\usepackage{multirow}
\usepackage{booktabs}
\usepackage{hyperref}

\usepackage{pifont}
\newcommand{\cmark}{\ding{51}} 
\newcommand{\xmark}{\ding{55}} 

\usepackage[capitalise]{cleveref}
\def\BibTeX{{\rm B\kern-.05em{\sc i\kern-.025em b}\kern-.08em
    T\kern-.1667em\lower.7ex\hbox{E}\kern-.125emX}}
\usepackage{balance}
\makeatletter
\renewcommand{\maketag@@@}[1]{\hbox{\m@th\normalsize\normalfont#1}}%
\makeatother
\makenomenclature
\usepackage{etoolbox}
\renewcommand\nomgroup[1]{%
  \item[\bfseries 
  \ifstrequal{#1}{C}{Variables}{%
  \ifstrequal{#1}{B}{Parameters}{%
  \ifstrequal{#1}{A}{Acronyms}{}}}%
]}
\begin{document}

\title{From Multi-Modal Paths to Executable Trajectories: A Trajectory Planning Framework for 4WIS Robots}


\author{{Runjiao Bao~\orcidlink{0009-0009-7999-8638}}, {Lin Zhang~\orcidlink{0000-0001-7281-8790}}, {Yongkang Xu~\orcidlink{0000-0003-0217-9035}}, {Shoukun Wang~\orcidlink{0000-0002-2433-6655}}

\thanks{This work was supported by the National Natural Science Foundation of China under Grant 62473044. \textit{(Corresponding author: Shoukun Wang.)}

Runjiao Bao is with the Department of Mechanical and Automation Engineering, The Chinese University of Hong Kong, Shatin, N.T., Hong Kong SAR, China (e-mail:runjiao.bao@link.cuhk.edu.hk).
Lin Zhang and Shoukun Wang are with the School of Automation, Beijing Institute of Technology, Beijing 100081, China (e-mail: bit.zhanglin@bit.edu.cn; bitwsk@bit.edu.cn).
Yongkang Xu is with the Institute of Intelligence Technology and Robotic Systems, Shenzhen Research Institute of Nankai University, Shenzhen 518000, China (e-mail: xuyk@nankai.edu.cn).
}}

\maketitle

\begin{abstract}
Four-wheel independent steering (4WIS) mobile robots support multiple motion modes, offering high maneuverability in narrow and complex environments. However, existing planning methods often fail to fully exploit these capabilities, leading to suboptimal trajectory quality. To address this limitation, this paper proposes a multi-modal global trajectory planning framework that couples mode-augmented front-end search with mode-consistent segment-wise trajectory optimization. In the front-end stage, Hybrid A* is extended to a four-dimensional state space incorporating motion modes, while mode-switching-aware cost and heuristic functions embed mode decisions into the global search process. Multi-modal Reeds-Shepp curves and an intelligent terminal connection strategy are further designed to improve search efficiency. In the back-end stage, a segment-wise trajectory optimization framework based on an improved iterative safe corridor scheme is developed to convert discrete multi-modal paths into smooth, kinematically feasible trajectories with stationary mode transitions. Experimental results show that the proposed method achieves the best overall performance in safety, arrival time, terminal accuracy and computation time. Real-world experiments on a physical 4WIS robot further validate the practical effectiveness and executability of the generated trajectories, providing a flexible and high-performance solution for multi-modal mobile robot trajectory planning.
\end{abstract}

\begin{IEEEkeywords}
4WIS mobile robot, Multi-modal trajectory, Search-based path planning, Segmented trajectory optimization
\end{IEEEkeywords}

\section{Introduction}
In recent years, driven by continuous advances in automation and intelligent technologies, mobile robots have gradually moved beyond fixed scenarios in traditional manufacturing and are increasingly deployed in complex and unstructured environments, including industrial inspection and disaster response under constrained or extreme conditions \cite{1}. In these tasks, conventional steering mechanisms often suffer from large turning radii and limited degrees of freedom, making it difficult to meet high-maneuverability requirements and thus promoting the evolution of mobile robot chassis structures \cite{3}. Four-wheel independent steering (4WIS) systems maintain good terrain adaptability and payload capacity while enabling flexible motion pattern switching through multiple steering modes. Fundamentally, different steering modes impose distinct wheel constraint conditions, thereby leading to differentiated kinematic characteristics of the system \cite{4}.

Although substantial progress has been made in trajectory tracking control and kinematic modeling of 4WIS robots \cite{5,6}, their global navigation and trajectory planning in complex environments remain insufficiently explored. Existing planning methods typically generate paths or trajectories in a mode-agnostic manner, while motion mode selection is deferred to the execution stage \cite{8}. As a result, motion modes are not explicitly coordinated at the global planning level, increasing execution uncertainty. In particular, mode switching during motion may introduce transient instability and reduce reliability \cite{9}, motivating the restriction of mode transitions to near-zero velocity states. Moreover, the few studies that explicitly consider multi-modal motion in global planning usually produce only time-parameter-free discrete paths rather than executable trajectories \cite{10}. Therefore, explicitly modeling multi-modal motion characteristics within a global planning framework and bridging multi-modal path generation with executable trajectory construction remain key challenges for 4WIS robot navigation.

To address these challenges, this paper proposes a multi-modal global trajectory planning framework for 4WIS robots. Instead of treating motion-mode selection as a post-processing or execution-stage decision, the proposed method embeds mode decisions into both global path search and continuous trajectory optimization. Extensive simulations and real-world experiments validate the effectiveness of the proposed method. The main contributions are as follows:
\begin{enumerate}[1.]
\item A mode-augmented Hybrid A* search formulation is developed for 4WIS global planning. By extending the search state, the proposed planner treats the motion mode as an explicit discrete planning variable and evaluates intra-mode motion primitives and inter-mode switching candidates within a unified search process.
\item A mode-switching-aware front-end evaluation mechanism is proposed to coordinate motion modes during global planning. Mode-dependent terminal connections, mode-switching costs, and the corresponding heuristic function are jointly incorporated into the node score, so that mode selection is guided by the global search cost rather than deferred to the trajectory execution stage.
\item A mode-consistent segment-wise trajectory optimization formulation is developed to transform discrete multi-modal paths into executable trajectories. The mode labels and switching nodes generated by the front-end search are inherited by the back-end optimizer as segment modes and switching constraints, where mode-specific dynamics, shared switching states, and near-zero switching velocities are imposed to ensure physical executability.
\end{enumerate}

\section{Related Works}

\subsection{Mobile Robots Motion Planning}

Motion planning methods for mobile robots can be broadly classified into four categories: search-based methods, sampling-based methods, optimization-based methods, and learning-based methods. Search-based methods discretize the state space into graphs or grids, providing strong global optimality and straightforward implementation, but often neglect vehicle kinematic constraints, leading to infeasible paths \cite{A*}. Hybrid A* alleviates this issue by incorporating motion primitives and has therefore been widely adopted for ground robot global planning \cite{HA*}. Sampling-based methods construct feasible paths via random sampling \cite{RRT*,KRRT*}, offering good scalability in high-dimensional spaces and under complex constraints. However, their probabilistic convergence makes it difficult to guarantee solution quality within limited computation time.

Optimization-based methods formulate trajectory generation as continuous constrained optimization problems, enabling smooth and high-quality trajectories \cite{opt1,opt2}. Their performance, however, strongly depends on the initial solution and constraint formulation, and strict constraints may lead to convergence issues or solver failure \cite{hard}. As a result, soft constraints are commonly adopted to improve numerical robustness \cite{soft}. Learning-based methods have attracted increasing attention in recent years. Reinforcement learning optimizes control policies through interaction with the environment \cite{DRL-VO,RMRL}, while supervised learning maps perception to control using expert demonstrations \cite{NeuPAN,RDNN}. Learning-based methods can also be combined with traditional optimization modules, where learned cost-to-go estimates are used to guide reference trajectory generation and facilitate constrained optimization \cite{add1}. Despite their flexibility, such methods typically suffer from high training cost, limited generalization, and sim-to-real transfer challenges.

Overall, different paradigms are suited to different roles: search-based and sampling-based methods are commonly used as front-end global planners, optimization-based methods serve as back-end trajectory refinement or local planners, while learning-based approaches are primarily applied to local obstacle avoidance and short-horizon decision-making.

\subsection{4WIS Robots Motion Planning}

Compared with conventional mobile robots, 4WIS robots are characterized by multiple motion modes and mode switching capability. Their control inputs combine continuous variables with discrete mode selections, making the system inherently hybrid. Existing studies on 4WIS robots have largely focused on single-mode control strategies, including kinematic and dynamic modeling \cite{RW1} and trajectory tracking or motion control methods tailored to specific steering configurations \cite{RW2,RW3}, while motion mode switching according to task requirements is generally not considered.

Studies that address motion mode switching can be broadly categorized into two classes: execution-stage mode decision and planning-stage mode modeling. Most prior work belongs to the former category, where motion modes are determined during trajectory execution based on a predefined reference trajectory. Nguyen et al. \cite{9} formulated a mixed-integer model predictive control scheme to handle non-stationary switching among multiple steering modes, Wang et al. \cite{8,RW4} addressed mode decision and tracking using hierarchical fuzzy control, and Bao et al. \cite{RW5} employed hierarchical deep reinforcement learning for joint mode selection and tracking. However, since multi-modal motion characteristics are not explicitly considered during planning, these approaches have limited ability to exploit the motion potential of 4WIS robots at the global level. In contrast, only a few studies have incorporated multi-modal motion characteristics into global planning. Zhang et al. \cite{RW6} proposed a hierarchical scheduling framework combining hybrid meta-heuristic optimization with kinematics-aware conflict path planning, while Chang et al. \cite{10} designed a hierarchical path planner that explicitly considers multiple motion modes. Nevertheless, these methods typically generate only coarse geometric paths and do not further construct executable trajectories that satisfy kinematic and dynamic constraints. Overall, a clear research gap remains between multi-modal path planning and dynamically feasible trajectory generation for 4WIS robots.

\section{Preliminaries}

\subsection{Trajectory Planning Problem Statement}

In the global trajectory planning task, we consider a bounded planar workspace $W \subset \mathbb{R}^2$ containing static obstacles represented by a closed set $\mathcal{O} \subset W$. The obstacle-free region is defined as $W_{\text{free}} = W \setminus \mathcal{O}$. The robot configuration is defined as $\mathbf{s} = (x, y, \theta)^T \in \mathcal{S} = \mathbb{R}^2 \times \mathbb{S}^1$, 
where $(x, y)$ denotes the position and $\theta \in [0, 2\pi)$ the heading angle. 
Let $R(\mathbf{s}) \subset W_{\text{free}}$ denote the region occupied by the robot at configuration $\mathbf{s}$, the collision-free configuration space is: 
\begin{equation}
\mathcal{S}_{\text{free}} =
\left\{
\mathbf{s} \in \mathcal{S} \mid R(\mathbf{s}) \subseteq W_{\text{free}}
\right\}.
\end{equation}

Given an initial configuration $\mathbf{s}_{\text{init}} \in \mathcal{S}_{\text{free}}$ and a goal configuration $\mathbf{s}_{\text{goal}} \in \mathcal{S}_{\text{free}}$, 
the planning problem aims to compute a feasible trajectory $\gamma : [0,1] \rightarrow \mathcal{S}_{\text{free}}$ satisfying $\gamma(0) = \mathbf{s}_{\text{init}}$ and $\gamma(1) = \mathbf{s}_{\text{goal}}$.

In practical implementations, the continuous trajectory is commonly approximated by a discrete sequence of \(\mathcal{N}+1\) waypoints:
\begin{equation}
\mathcal{T} = \{\mathbf{s}_i\}_{i=0}^{\mathcal{N}},
\end{equation}
where \(\mathbf{s}_0 = \mathbf{s}_{\text{init}}\) and \(\mathbf{s}_{\mathcal{N}} = \mathbf{s}_{\text{goal}}\).
A valid path must satisfy both kinematic feasibility and collision avoidance constraints, where the transitions between adjacent waypoints must comply with the robot’s limits, such as steering and motion continuity requirements, and all waypoints must remain entirely within the collision-free space $\mathcal{S}_{\text{free}}$.

\subsection{4WIS Robots Motion Modes}

\begin{figure}[t]
\centering
    \includegraphics[width=0.7\columnwidth]{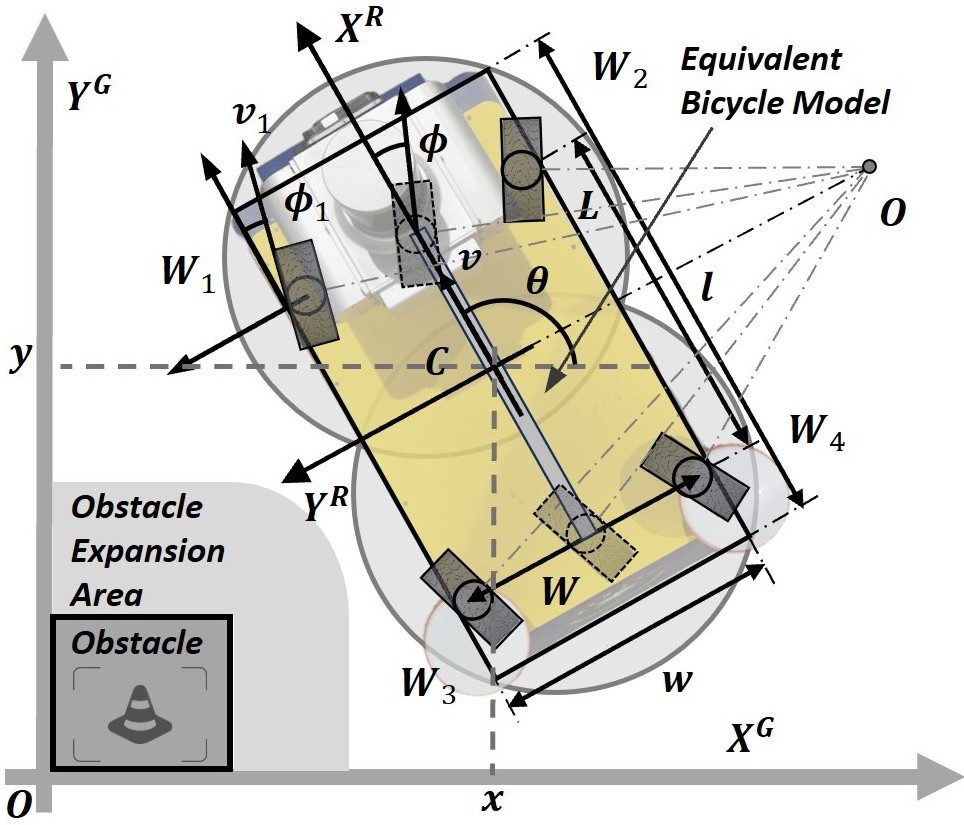}
\caption{Configuration of the 4WIS robot.}
\label{fig1}
\vspace{-5pt}
\end{figure}

The 4WIS system supports three classical motion modes: Ackermann steering, lateral steering, and parallel movement. As shown in \cref{fig1}, \(C\) denotes the robot center point, \(L\) denotes the wheelbase, \(W\) denotes the track width, \(l\) and \(w\) denote the overall body length and width, respectively. 


Although the 4WIS robot has a complex steering
kinematic structure, it can be simplified using a classical equivalent
bicycle model when tire slip is negligible. In this model, the coordinated
rolling constraints of the four physical wheels are approximated by
virtual wheels on the effective centerline of motion. Thus, \(\phi\)
denotes the steering angle of the virtual wheel under the corresponding
motion mode, rather than that of any individual physical wheel, while
\(v\) denotes the velocity of the robot center point \(C\), rather than
the rolling speed of any wheel. At the control layer, \(v\) and \(\phi\)
are mapped to the steering angles and driving velocities of the four
physical wheels based on the double-Ackermann steering principle \cite{add-1}.

Based on the above equivalent bicycle representation, the complete system
state and control input are defined as
\[
\mathbf{x} =
\big[x, y, v, \theta, \phi\big]^{\mathsf{T}}, \qquad
\mathbf{u} =
\big[a, \omega\big]^{\mathsf{T}},
\]
where \((x,y)\) denotes the position of the robot center point \(C\),
\(\theta\) denotes the robot heading angle, \(v\) denotes the signed
velocity of \(C\), and \(\phi\) denotes the equivalent steering angle.
The control variables \(a\) and \(\omega\) denote the acceleration of
\(v\) and the steering rate of \(\phi\), respectively.

 Assume the discrete time step is $\Delta t$, The discrete kinematic model of the system is defined as:
\begin{equation}
\label{motion1}
\mathbf{x}_{i+1} = \mathbf{x}_i+\big(f_m(\mathbf{x}_i)+B\mathbf{u}_i\big)\Delta t,
\end{equation}
where $f_m(\cdot)$ denotes the body kinematic function associated with motion mode $m$ and \(i\) denotes the discrete time index.
The control input matrix $B$ remains identical across all motion modes and is given by:
\begin{equation}
B =
\big[
[0,0],
[0,0],
[1,0],
[0,0],
[0,1]
\big]^{\mathsf{T}}\!.
\end{equation}

In Ackermann mode, the equivalent bicycle model is constructed along the longitudinal axis of the robot. Under the no-slip rolling constraint, the normal lines of the wheel rolling directions intersect at a common instantaneous center of rotation (ICR). In this mode, the ICR is
constrained to lie on the \(Y^R\)-axis of the body-fixed frame, and the velocity of the robot center \(C\) is aligned with the longitudinal axis \(X^R\). This mode is suitable for conventional driving and smooth turning in relatively open spaces. Accordingly, the body kinematic function for the Ackermann mode is given by:
\begin{equation}
\label{motion2}
f_{\mathrm{ack}}(\mathbf{x}) =
\big[v\cos\theta, v\sin\theta, 0, 2v\tan\phi/L, 0 \big]^{\mathsf{T}}\!.
\end{equation}

Lateral steering can be regarded as an Ackermann-like motion in which the lateral axis \(Y^R\) serves as the effective forward direction. In this mode, the equivalent bicycle model is constructed along the lateral axis of the robot. The ICR is constrained to lie on the \(X^R\)-axis of the body-fixed frame, and the velocity of the robot center \(C\) is aligned with the lateral axis \(Y^R\). This mode is particularly suitable for structured parking scenarios and lateral pose adjustment. By preserving ICR consistency and exchanging $L$ and $W$, the corresponding body kinematic model can be obtained as:
\begin{equation}
\label{motion3}
f_{\mathrm{lat}}(\mathbf{x}) =
\big[-v\sin\theta,v\cos\theta,0,2v\tan\phi/W,0\big]^{\mathsf{T}}\!.
\end{equation}

In parallel movement, all wheels have identical steering angles and wheel velocities. Accordingly, the ICR is located at infinity, and the robot translates with a constant heading. This mode is suitable for fixed-orientation obstacle avoidance and precise docking or alignment tasks, with the velocity of the robot center \(C\) aligned with the common wheel direction:
\begin{equation}
\label{motion4}
f_{\mathrm{par}}(\mathbf{x}) =
\big[v\cos(\theta+\phi),v\sin(\theta+\phi),0,0,0\big]^{\mathsf{T}}\!.
\end{equation}

\subsection{Collision Model}
\label{collision}
To simplify geometric modeling and improve computational efficiency, we approximate the robot footprint using a two-circle representation. Specifically, two circles with identical radius are placed along the robot longitudinal axis to cover the front and rear parts of the body. Given the robot state $(x,y,\theta)$, the centers of the front and rear circles in the global frame are obtained via a rigid-body transform:
\begin{equation}
\begin{aligned}
O_f^x = x + l\cos\theta/4,\qquad O_f^y = y + l\sin\theta/4,\\
O_r^x = x - l\cos\theta/4,\qquad O_r^y  = y - l\sin\theta/4,
\end{aligned}
\label{circle}
\end{equation}
and the circle radius $r_b$ is defined as follows:
\begin{equation}
r_b = \sqrt{(l/4)^2+(w/2)^2},
\end{equation}
where $(O_f^x,O_f^y)$ and $(O_r^x,O_r^y)$ denote the front and rear circle centers. For fast collision checking, we employ an obstacle inflation scheme in which the obstacle set \(\mathcal{O}\) is dilated by the closed disk \(\mathcal{B}(r_c)\) through the Minkowski sum \(\oplus\), where \(r_c=r_b+r_{\mathrm{safe}}\) denotes the inflation radius with an optional safety margin \(r_{\mathrm{safe}}\ge 0\). Under this representation, collision avoidance reduces to requiring that both circle centers lie outside the inflated obstacle region:
\begin{equation}
(O_f^x,O_f^y)\notin \mathcal{O}\oplus\mathcal{B}(r_c),\quad
(O_r^x,O_r^y)\notin \mathcal{O}\oplus\mathcal{B}(r_c).
\end{equation}
This compact formulation enables efficient collision checking while conservatively accounting for the robot footprint, with \(r_{\mathrm{safe}}\) set to 0.02 m in all experiments of this paper.

\begin{figure*}[t]
\centering
\includegraphics[width=0.92\textwidth]{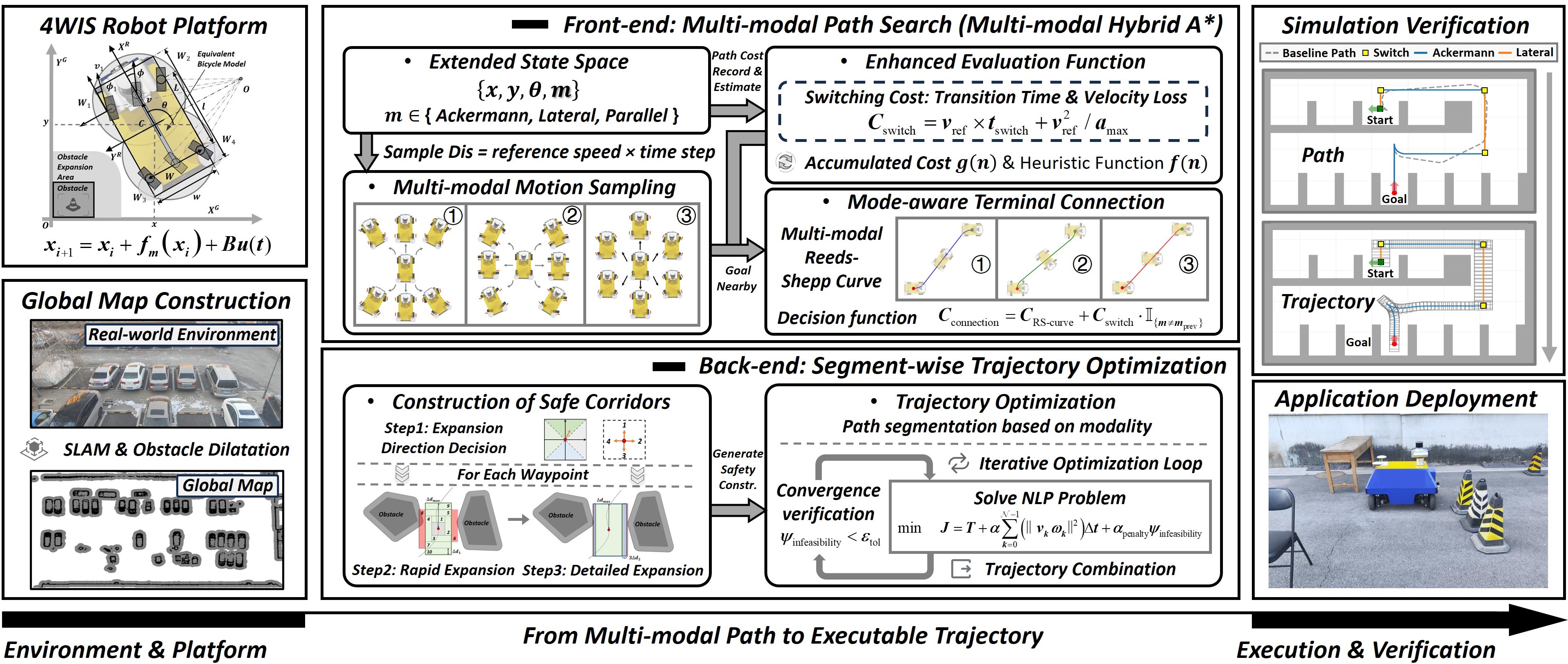}
\caption{The overall architecture diagram of the 4WIS multi-modal global trajectory planning framework.}
\label{fig2}
\vspace{-5pt}
\end{figure*}

\section{Methodology}

The proposed multi-modal global trajectory planning framework for 4WIS robots is depicted in \cref{fig2}. It integrates front-end path generation (\cref{Front}) with back-end trajectory optimization (\cref{Back}) to achieve collision-free and kinematically feasible navigation in complex environments.

\subsection{Front-end Path Generation}
\label{Front}

\subsubsection{Multi-Modal State Representation}

Hybrid A* augments grid-based A* search with continuous motion-primitive expansions, enabling it to directly generate collision-free and kinematically feasible paths. We adopt Hybrid A* as the baseline primarily because its motion primitives, heuristic design, and terminal-connection module are highly modular, which facilitates integrating the multi-modal maneuverability of 4WIS into a unified search framework. The node evaluation function is defined as:
\begin{equation}
f(n) = g(n) + h(n),
\end{equation}
where $g(n)$ denotes the accumulated cost and $h(n)$ is the heuristic estimate. When the search approaches the goal region, the algorithm typically attempts a terminal connection using Reeds–Shepp (RS) curves \cite{RS} to accelerate convergence. If the connecting trajectory is collision-free, a solution is returned immediately, otherwise the search continues with regular expansions.

Traditional Hybrid A* algorithms are designed for robots with fixed motion patterns, typically based on Ackermann steering models, and cannot fully exploit the multi-modal motion advantages of four-wheel independent steering systems. To overcome the limitations of traditional algorithms, this research enhances state representation and search strategies to achieve unified modeling of multiple motion modes. The core idea is to extend the original three-dimensional state space $(x, y, \theta)$ into a four-dimensional hybrid state space $(x, y, \theta, m)$, where the motion mode variable $m$ is explicitly modeled as the fourth dimension of the state. Specifically, $m \in \{1, 2, 3\}$ corresponds to Ackermann steering, lateral steering and parallel movement. In the parallel movement mode, the robot heading $\theta$ remains unchanged during primitive propagation, and the common wheel angle $\phi$ is used as the direction variable for state propagation.

To make the motion-primitive length consistent across different modes, we introduce a reference velocity and sampling time mechanism to replace the traditional fixed forward iteration distance method. By defining reference velocity $v_{\text{ref}}$ and iteration time $\Delta t$ as constants, the forward iteration distance is dynamically calculated as $\Delta s = v_{\text{ref}} \times \Delta t$. This improvement lays the foundation for subsequent multi-modal RS curve expansion and mode switching cost quantification.

\begin{figure}[t]
\centering
\includegraphics[width=\columnwidth]{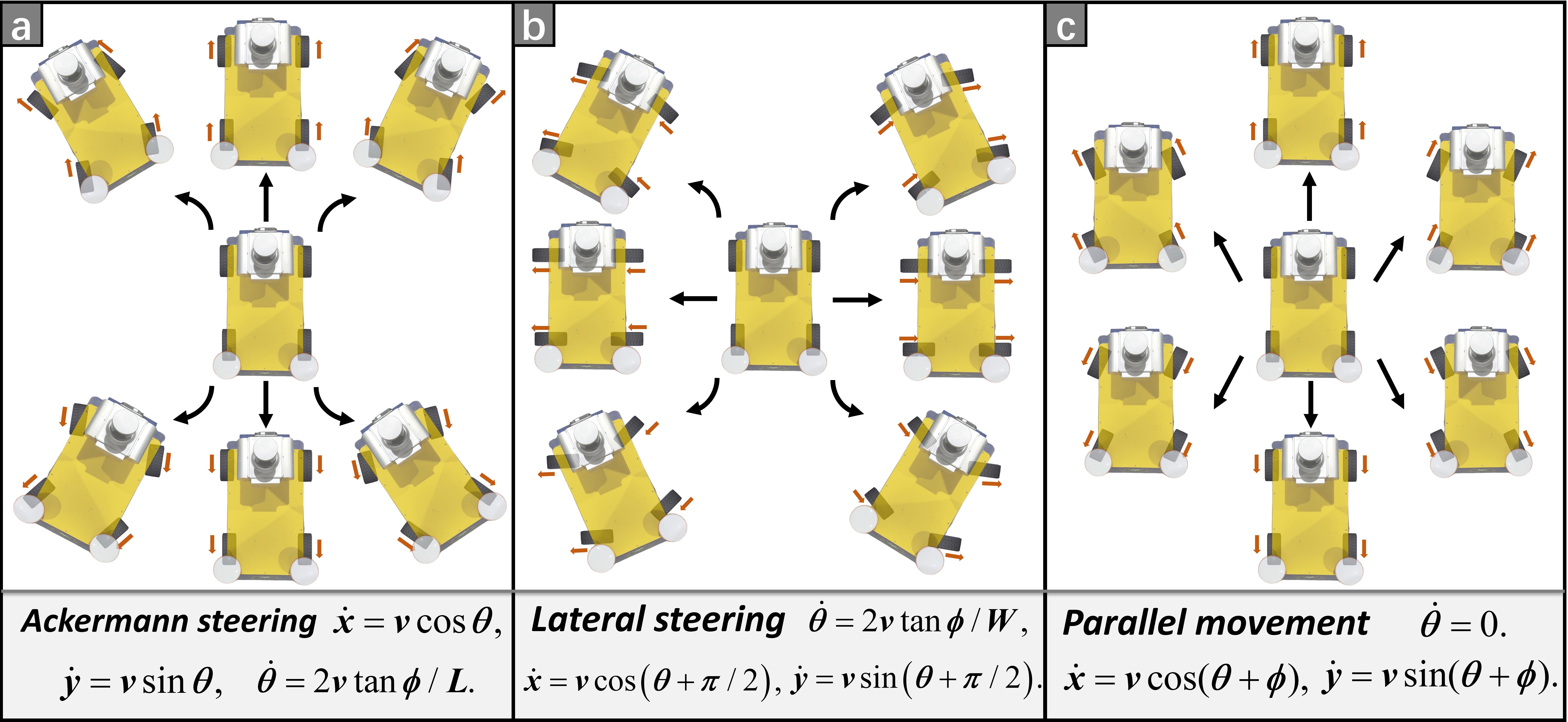}
\caption{Multi-modal motion sampling. (a) Ackermann steering. (b) Lateral steering. (c) Parallel movement.}
\label{fig3}
\vspace{-5pt}
\end{figure}

Within the extended 4D state space, the expansion procedure comprises two types of operations: intra-mode expansion and inter-mode expansion. Intra-mode expansion keeps the current motion mode unchanged and generates successor nodes using mode-specific motion primitives. For Ackermann steering and lateral steering, the robot heading \(\theta\) is updated according to the corresponding yaw-rate model. The equivalent steering angle is directly sampled as:
\begin{equation}
\phi \in \{-\phi_{\max},0,\phi_{\max}\}.
\end{equation}
Both forward and backward driving directions are considered. For parallel movement, the robot heading \(\theta\) remains unchanged, while the translational direction is determined by the common wheel angle \(\phi\). Therefore, the steering-angle increment is sampled and the common wheel angle is updated as:
\begin{equation}
\Delta\phi_{\mathrm{par}}
\in
\{-\dot{\phi}_{\max}\Delta t,0,\dot{\phi}_{\max}\Delta t\},
\phi_{i+1}
=
\phi_i+\Delta\phi_{\mathrm{par}},
\end{equation}
where $\dot{\phi}_{\text{max}}$ denotes the maximum steering rate. The updated \(\phi_{i+1}\) is then used to determine the translational direction of the parallel-motion primitive. Except for the parallel movement mode, \(\phi\) is not included as a dimension of the closed-list search state. Instead, it is retained as a primitive-associated auxiliary attribute and is used only for successor generation and cost evaluation. The mode-specific sampling strategies are illustrated in \cref{fig3}, and the corresponding constraint formulations are provided in \cref{motion2,motion3,motion4}.

 In contrast, inter-mode expansion inserts a switching node at the current searched pose while changing only the motion-mode label. This operation is defined at the front-end path-search level, indicating that the robot is expected to stop at this waypoint and perform mode reconfiguration rather than switch modes instantaneously at a nonzero velocity. The corresponding stopping and switching effects are considered through the mode-switching cost \(C_{\mathrm{switch}}\), while the continuous deceleration and acceleration processes are generated by the back-end segment-wise trajectory optimization.

\subsubsection{Multi-Modal Reeds-Shepp Curves}

To accommodate the multi-modal motion of a 4WIS robot, we extend the RS curve formulation by introducing mode-dependent curvature bounds. In standard RS curve construction, motion primitives are parameterized by a constant upper bound on curvature derived from the vehicle kinematics. For 4WIS systems, the curvature-generation mechanism varies fundamentally across modes. While the Ackermann and lateral steering modes admit geometric curvature limits determined by an effective wheelbase and the maximum steering angle, the parallel translation mode yields a curvature bound explicitly coupled with translational speed. Accordingly, for mode $m$ we define:
\begin{equation}
\kappa_{\text{max}}^{(m)} =
\begin{cases}
2\tan\phi_{\text{max}}/L, &m=1,\\
2\tan\phi_{\text{max}}/W, &m=2,\\
\dot{\phi}_{\text{max}}/v_{\text{ref}}, &m=3,
\end{cases}
\end{equation}
where $\phi_{\max}$ is the maximum steering angle, and $\kappa$ is the path curvature. For the lateral steering mode, we reuse the standard RS construction via an equivalent coordinate transform that swaps the effective forward axis from longitudinal to lateral and treats the track width $W$ as the equivalent wheelbase, the resulting curve is then mapped back to the original frame. The parallel translation mode is treated as a special case and is activated only when the heading error is below a prescribed tolerance, to avoid unnecessary orientation mismatch.

\subsubsection{Cost and Heuristic Functions}

Based on the proposed multi-modal RS curves, we unify the formulation of the accumulated cost and the heuristic in Hybrid A* such that the additional cost induced by mode transitions is explicitly represented, thereby alleviating the systematic underestimation of conventional heuristics under complex maneuvers. Specifically, we model a single mode switch as an equivalent process consisting of velocity reconfiguration and switching latency, with its cost defined as:
\begin{equation}
C_{\text{switch}} = v_{\text{ref}} \times t_{\text{switch}} + v_\text{ref}^2/a_{\text{max}},
\end{equation}
where $t_{\text{switch}}$ denotes the preset switching time and $a_{\max}$ is the maximum acceleration. When the mode transition occurs at the initial node, the system starts at rest and thus incurs only the acceleration phase, leading to a halved velocity-related term. With $C_{\text{switch}}$, the extended accumulated cost is updated by:
\begin{align}
\!\!& g(n) = g_{\text{prev}} + \Delta s 
       + C_{\text{reverse}} \!\cdot\! \mathbb{I}_{\{d = -1\}} 
       + C_{\text{steer}} \!\cdot\! |\phi| \notag \\
       \!\!+  \ & C_{\Delta \text{steer}} \!\cdot\! |\Delta \phi| +  C_{\text{dir}} \!\cdot\! \mathbb{I}_{\{d \neq d_{\text{prev}}\}} + C_{\text{switch}} \!\cdot\! \mathbb{I}_{\{m \neq m_{\text{prev}}\}}.
\label{eq:front_cost}
\end{align}
where $g_{\text{prev}}$ is the accumulated cost of the parent node, $C$ are weighting coefficients, $d$ indicates the driving direction, and $\mathbb{I}$ is the indicator function. At mode-switching nodes, only $C_{\text{switch}}$ is retained and all other terms are set to zero. 

To accommodate the multi-modal setting, we further construct a heuristic based on the multi-modal RS connection cost, this design provides more accurate cost estimates while inheriting the admissibility of the Hybrid A* algorithm:
\begin{align}
\!\! h(n) = \max\{h_{\text{euc}}, \min_{m \in M} [h_{\text{RS}}^{(m)}+ C_{\text{switch}} \cdot \mathbb{I}_{\{m \neq m_{\text{prev}}\}}]\},\!\!
\end{align}
where $h_{\text{euc}}$ is the heuristic term based on Euclidean distance, $h_{\text{RS}}^{(m)}$ is the RS curve heuristic term based on corresponding motion mode, and $M$ is the set of all available motion modes. 

Based on the above cost and heuristic formulation, no separate rule-based mode pruning is introduced. Feasible successors of different motion modes are evaluated by the unified score \(f(n)=g(n)+h(n)\), where both the accumulated cost and heuristic term include the corresponding mode-switching penalty. Therefore, lower-cost modes are naturally prioritized by the open-list ordering, while other feasible modes remain available for later expansion.

\subsubsection{Terminal Connection Strategy}

When the search approaches the goal, the algorithm attempts direct terminal connections using multi-modal RS curves. Candidate connections include RS paths generated in the current mode and those obtained after switching to alternative modes, with the corresponding switching costs taken into account. Among all feasible candidates, the connection with the minimum cost is selected as the terminal path. The total connection cost is:
\begin{equation}
C_{\text{connection}} = C_{\text{RS-curve}} + C_{\text{switch}} \cdot \mathbb{I}_{\{m \neq m_{\text{prev}}\}},
\end{equation}
where $C_{\text{RS-curve}}$ is the cost of the RS curve. This terminal connection strategy ensures both kinematic feasibility of the path and optimizes connection efficiency through mode selection.

\begin{algorithm}[t]
\caption{Front-end multi-modal Path Generation}
\label{alg:multimodal_hybrid_astar}
\begin{algorithmic}[1]
\State Initialize open list with start node $(x_0, y_0, \theta_0, m_0)$
\State Initialize closed list as empty
\While{open list is not empty}
    \State $n \leftarrow$ node with minimum $f(n)$ from open list
    \If{$n$ is close to goal}
        \State Compute all multi-modal RS curves to goal
        \State Sort RS curves by ascending connection cost 
        \For{each RS curve in sorted order}
            \If{RS curve is collision-free}
                \State \Return path 
            \EndIf
        \EndFor
    \EndIf
    \State Move $n$ to closed list
    \For{each available motion mode $m'$}
        \If{$m' = m_{\text{current}}$}
            \State Generate intra-modal successors
        \Else
            \State Generate inter-modal successors (switch)
        \EndIf
        \State Compute $g(n')$ with mode switching cost
        \State Compute multi-modal heuristic $h(n')$
        \If{$n'$ is collision-free and not in closed list}
            \State Add $n'$ to open list
        \EndIf
    \EndFor
\EndWhile
\State \Return failure
\end{algorithmic}
\end{algorithm}

\subsection{Back-end Path Optimization}
\label{Back}

\subsubsection{Construction of Safe Corridors}

To transform the discrete multi-modal path generated by the front-end planner into a continuous and executable trajectory, the back-end optimization stage introduces a local safe-corridor-based constraint formulation. By constructing a sequence of convex feasible regions in the vicinity of the reference path, the safe corridor provides a restricted yet connected search space for continuous trajectory optimization. 

Specifically, the discrete path produced by the front-end planner is first resampled with a prescribed spatial interval $\mathcal{D}$, and the orientation angles along the path are processed to ensure continuity, resulting in a geometrically well-distributed sequence of $\mathcal{N}+1$ waypoints. Centered at each resampled waypoint, a local convex safety region is constructed under obstacle constraints, yielding a corridor sequence
$\mathcal{C}_{\text{seq}} = \{\mathcal{C}_i\}_{i=1}^{\mathcal{N}}$.

The constructed corridor satisfies the inclusion relation:
\begin{equation}
\mathcal{C}_{\text{seq}} \subseteq W \setminus (\mathcal{O} \oplus \mathcal{B}(r_c)),
\end{equation}
which guarantees collision-free feasibility within the corridor.In implementation, environment obstacles are first inflated according to the geometric approximation model described in \cref{collision}. 

Based on the pose information of each path point, the centers of the front and rear circles of the bi-disc robot model are computed, generating corresponding front-center and rear-center reference paths. Dynamic safety corridors are then synchronously constructed along these paths. For each discrete waypoint, an axis-aligned square region with side length $\Delta d_1$ is initialized at the waypoint center.

\begin{figure}[t]
\centering
\includegraphics[width=\columnwidth]{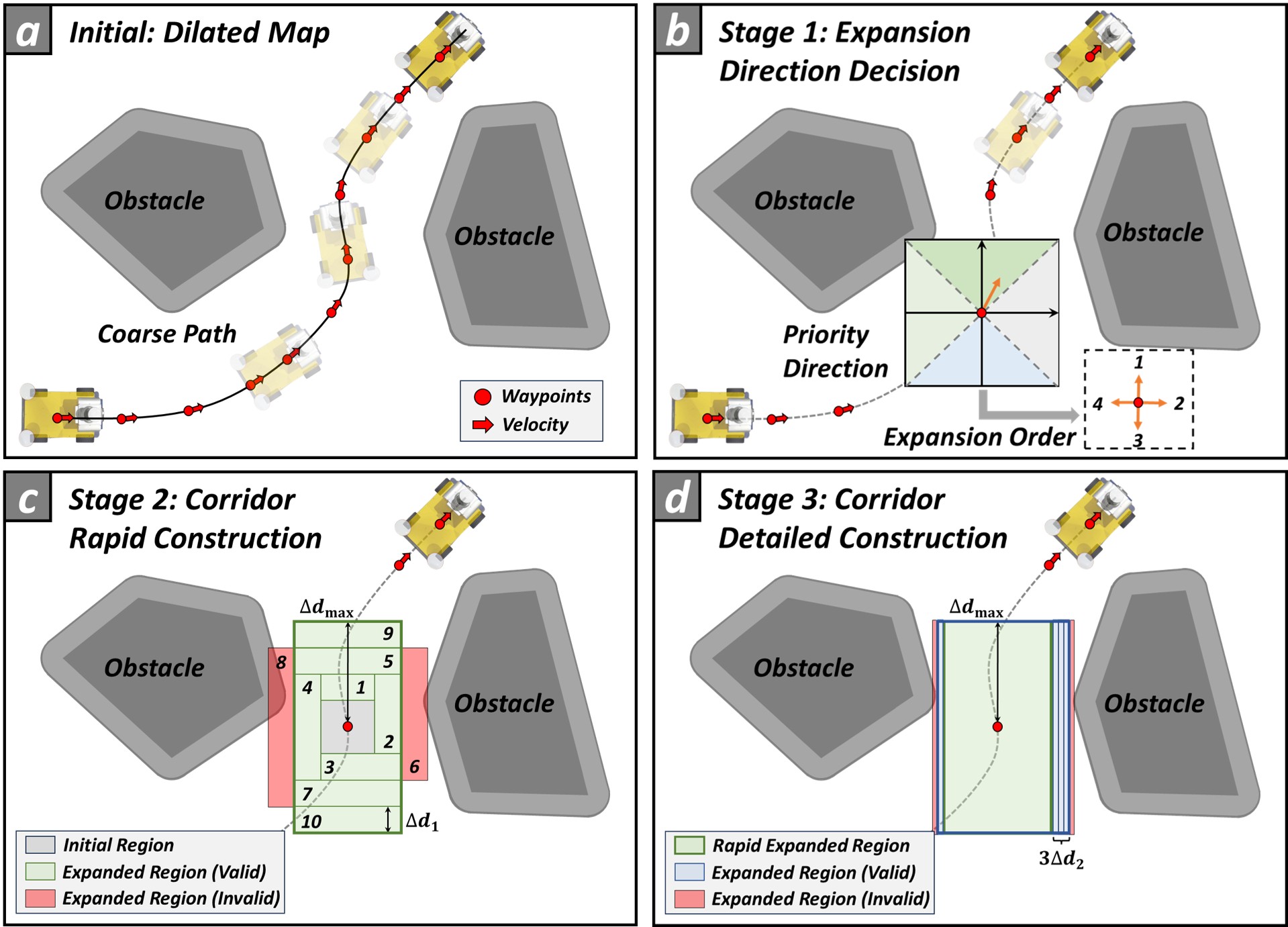}
\caption{Improved adaptive safe corridor construction process. (a) Obstacle dilatation. (b) Direction decision. (c) Rapid expansion. (d) Detailed expansion.}
\label{fig4} 
\vspace{-5pt}
\end{figure}

Unlike conventional approaches that expand the corridor cyclically along fixed orthogonal directions (i.e., $+x$, $+y$, $-x$, and $-y$), we introduce a motion-direction-guided adaptive expansion strategy. The planar workspace is partitioned into four areas according to the local velocity direction at each waypoint, which determines the priority expansion direction. The safety region is expanded along this direction with step size $\Delta d_1$, accompanied by immediate collision checking after each expansion. If no collision is detected, the expansion continues. Otherwise, the expansion step is reduced to $\Delta d_2$ for local refinement. The expansion process terminates when a collision occurs again or the maximum allowable expansion distance $d_{\max}$ is reached along the corresponding direction. The expansion process of the corridor is shown in \cref{fig4}.

The resulting safe corridors can be directly converted into linear box constraints for the optimal control problem. To facilitate numerical optimization, the continuous-time trajectory optimization problem is discretized over the finite horizon $[0,T]$ using $\mathcal{N}$ uniform time steps, yielding a time step size $\Delta t = T / \mathcal{N}$. Let the front and rear corridor bounds associated with the $k$-th waypoint be $\mathcal{C}_k = [x^{k}_{f\text{min}}, x^{k}_{f\text{max}}] \times [y^{k}_{f\text{min}}, y^{k}_{f\text{max}}], \tilde{\mathcal{C}}_k = [x^{k}_{r\text{min}}, x^{k}_{r\text{max}}] \times [y^{k}_{r\text{min}}, y^{k}_{r\text{max}}]$. By enforcing corridor constraints at discrete time instants $t_k = kT/ \mathcal{N}$, $k = 1,\ldots,\mathcal{N}$, the following box constraints are imposed:
\begin{equation}
\begin{aligned}
O_f^x(t_k) &\in [x^{k}_{f\text{min}}, x^{k}_{f\text{max}}], &
O_f^y(t_k) \in [y^{k}_{f\text{min}}, y^{k}_{f\text{max}}], \\
O_r^x(t_k) &\in [x^{k}_{r\text{min}}, x^{k}_{r\text{max}}], &
O_r^y(t_k) \in [y^{k}_{r\text{min}}, y^{k}_{r\text{max}}] .
\end{aligned}
\end{equation}
This formulation guarantees collision avoidance while effectively alleviating numerical difficulties caused by nonlinear constraints in cluttered environments. \cref{alg:corridor} presents the detailed procedure of the improved adaptive safe corridor generation method.

\begin{algorithm}[t]
\caption{Improved Adaptive Safe Corridor Generation}
\label{alg:corridor}
\begin{algorithmic}[1]
\Require {Resampled path point sequence \(\mathcal{P}\), obstacle set \(\mathcal{O}\), maximum expansion distance \(d_{\max}\)}
\Ensure Safe corridor sequence \(\mathcal{C}_{\mathrm{seq}}\)
\State \(\mathcal{C}_{\mathrm{seq}} \leftarrow \emptyset\)
\For{each path point \(p_i \in \mathcal{P}\)}
    \State \(\mathcal{C}_i \leftarrow \text{InitializeBox}(p_i, \Delta d_1)\)
    \State \(\mathbf{v}_i \leftarrow p_{i+1} - p_i\)
    \State \(\text{DirSorted} \leftarrow \text{ExpansionOrderDecision}(\mathbf{v}_i)\)
    \State \(d_{\mathrm{total}}[q], \text{step}[q], \text{blocked}[q] \leftarrow 0, \Delta d_1, \text{false}\)
    \While{\(\exists\, q \in \text{DirSorted} \;\text{s.t.}\; \text{blocked}[q] = \text{false}\)}
        \For{each expansion direction \(q \in \text{DirSorted}\)}
            \If{\(\text{blocked}[q]\)}
                \State \textbf{continue}
            \EndIf
            \State \(\mathcal{C}_{\mathrm{test}} \leftarrow \text{ExpandBoundary}(\mathcal{C}_i, q, \text{step}[q])\)
            \If{\(\text{IsCollision}(\mathcal{C}_{\mathrm{test}})\) \textbf{or} \(d_{\mathrm{total}}[q] \geq d_{\max}\)}
                \If{\(\text{step}[q] = \Delta d_1\)}
                    \State \(\text{step}[q] \leftarrow \Delta d_2\)
                \Else
                    \State \(\text{blocked}[q] \leftarrow \text{true}\)
                \EndIf
            \Else
                \State \(\mathcal{C}_i,d_{\mathrm{total}}[q] \leftarrow \mathcal{C}_{\mathrm{test}}, d_{\mathrm{total}}[q] + \text{step}[q]\)
            \EndIf
        \EndFor
    \EndWhile
    \State \(\mathcal{C}_{\mathrm{seq}}.\mathrm{push}(\mathcal{C}_i)\)
\EndFor
\State \Return \(\mathcal{C}_{\mathrm{seq}}\)
\end{algorithmic}
\end{algorithm}

\subsubsection{Iterative Optimization Framework}

After obtaining the front-end multi-modal reference path together with its associated safety corridors, the path is first divided into \(S\) trajectory segments according to the mode-switching points. These switching points are fixed by the front-end planning result and serve as boundary conditions between segments during back-end optimization. Each segment keeps a fixed
motion mode and is optimized over its corresponding time interval \(T_r\) under the associated mode-specific kinematic constraints. The initial duration of the \(r\)-th segment is assigned according to its reference length \(L_r\):
\begin{equation}
T_r^{0}=L_r/v_{\mathrm{ref}}, \qquad r=1,\ldots,S .
\end{equation}

For adjacent segments, the shared mode-switching point preserves the same geometric pose, and the velocity is constrained to zero:
\begin{equation}
\!\!\mathbf{p}^{(r)}(T_r)
=
\mathbf{p}^{(r+1)}(0)
=
\mathbf{q}_{r},
\
v^{(r)}(T_r)=v^{(r+1)}(0)=0,
\end{equation}
where
\(\mathbf{p}^{(r)}(t)=[x^{(r)}(t),y^{(r)}(t),\theta^{(r)}(t)]^{\mathsf{T}}\) denotes the geometric pose of the \(r\)-th segment, and \(\mathbf{q}_{r}\) denotes the geometric pose of \(r\)-th mode-switching waypoint. Thus, the position and orientation of each switching point remain unchanged during back-end optimization, while the continuous states and controls inside each segment are optimized, ensuring that every mode transition is executed at rest.

The final trajectory is obtained by concatenating all optimized segments in sequence, with a mode-reconfiguration interval \(t_{\mathrm{switch}}\) inserted at each switching point:
\begin{equation}
T_{\mathrm{total}}
=
\sum_{r=1}^{S}T_r+(S-1)t_{\mathrm{switch}} .
\end{equation}
This formulation ensures pose consistency at mode-switching points while explicitly accounting for the time required for stationary mode reconfiguration.

For each trajectory segment, the back-end trajectory generation is formulated as an independent nonlinear programming (NLP) problem. The discretized states and controls of this segment are indexed by \(k=0,\ldots,N_r\), with \(\Delta t_r=T_r/N_r\), and its motion mode is fixed according to the front-end planning result. To alleviate numerical infeasibility caused by strict equality constraints and geometric consistency constraints, part of the constraints are relaxed using soft penalty terms. The initial state and control trajectories are initialized from the front-end reference path through temporal resampling and kinematic interpolation, providing a feasible warm start for the NLP solver and improving convergence robustness.

The discrete kinematic constraints are incorporated via a penalty formulation, with the corresponding cost defined as:
\begin{equation}
J_{\text{dyn}}
= \sum_{k=0}^{\mathcal{N}-1}
\left\|
\mathbf{x}_{k+1}-\mathbf{x}_k-\big(f_{m_k}(\mathbf{x}_k)+B\mathbf{u}_k\big)\Delta t
\right\|^2,
\end{equation}
where \(m_k\) denotes the fixed motion mode of the current trajectory segment specified by the front-end path.

Meanwhile, to avoid conflicts between the safe-corridor constraints and the geometric model, the front and rear circle centers are introduced as independent optimization variables. Their rigid-body relationship with the vehicle pose is enforced through a geometric consistency penalty term:
\begin{equation}
J_{\text{geom}}
=\sum_{k=0}^{\mathcal{N}}
\Big(
\big\|\mathbf{p}^f_k-\bar{O}_f(\mathbf{x}_k)\big\|_2^2
+
\big\|\mathbf{p}^r_k-\bar{O}_r(\mathbf{x}_k)\big\|_2^2
\Big)\,\Delta t,
\end{equation}
where $\mathbf{p}^f_k = [O_f^x(t_k),\, O_f^y(t_k)]^{\mathsf{T}}$ and $\mathbf{p}^r_k = [O_r^x(t_k),\, O_r^y(t_k)]^{\mathsf{T}}$ denote the optimized front and rear circle center positions at the $k$-th time step, respectively. $\bar{O}_f(\mathbf{x}_k)$ and $\bar{O}_r(\mathbf{x}_k)$ represent their nominal positions determined by the rigid-body geometric relationship defined \cref{circle}.

By incorporating the two aforementioned soft constraint terms, an infeasibility metric is defined as $\psi_{\text{infeasibility}} = J_{\text{dyn}} + J_{\text{geom}}.$. Accordingly, the objective of the back-end trajectory optimization is formulated as a weighted combination of task completion time, comfort cost $\|v_k\,\omega_k\|^2$, and constraint violation severity:
\begin{align}
\min \quad & J= T+ \alpha \sum_{k=0}^{\mathcal{N}-1}\left(\|v_k\,\omega_k\|^2\right)\Delta t + \alpha_{\text{penalty}}\psi_{\text{infeasibility}} \notag \\
\text{s.t.}   \quad
& \mathbf{x}_0 = \mathbf{x}_{\text{init}},\ k=1,\ldots,\mathcal{N}, \notag \\
& x_{\mathcal{N}} = x_{\text{goal}}, \
  y_{\mathcal{N}} = y_{\text{goal}}, \
  v_{\mathcal{N}} = v_{\text{goal}}, \
  \phi_{\mathcal{N}} = \phi_{\text{goal}}, \notag \\
& \sin \theta_{\mathcal{N}} = \sin \theta_{\text{goal}}, \
  \cos \theta_{\mathcal{N}} = \cos \theta_{\text{goal}}, \notag \\
& O_f^x(\mathbf{x}_k) \in [x^{k}_{f\text{min}}, x^{k}_{f\text{max}}],\ 
  O_f^y(\mathbf{x}_k) \in [y^{k}_{f\text{min}}, y^{k}_{f\text{max}}], \notag \\ 
& O_r^x(\mathbf{x}_k) \in [x^{k}_{r\text{min}}, x^{k}_{r\text{max}}],\ 
  O_r^y(\mathbf{x}_k) \in [y^{k}_{r\text{min}}, y^{k}_{r\text{max}}], \notag \\
& \mathbf{x}_{\text{min}} \le \mathbf{x}_k \le \mathbf{x}_{\text{max}}, \
  \mathbf{u}_{\text{min}} \le \mathbf{u}_k \le \mathbf{u}_{\text{max}},
\label{19}
\end{align}
where $\alpha$ is the weight factor and $\alpha_{\text{penalty}}$ is the penalty factor for the infeasibility metric. To prevent excessive violations of kinematic and geometric constraints, the infeasibility metric $\psi_{\text{infeasibility}}$ is employed as one of the termination criteria for the iterative process. When this metric falls below a threshold $\varepsilon_{\text{tol}}$ or the maximum iterations $\mathcal{I}$ is reached, the algorithm terminates and outputs the current trajectory. Otherwise, a new safe corridor is reconstructed based on the current solution, and the optimized state, control, and time variables from the previous iteration are used as the initial guess for the next NLP optimization. The optimization then proceeds iteratively until the termination condition is satisfied.

\section{Experiments}

\subsection{Experimental Setup}

To validate the proposed 4WIS trajectory planning framework, extensive simulation experiments were conducted on a computing platform equipped with an Intel i7-13700H CPU and an NVIDIA RTX 4060 GPU. The back-end trajectory optimization problem was formulated as a NLP problem using CasADi \cite{casadi} and solved with the IPOPT \cite{ipopt} solver. The key geometric, kinematic, and planning-related parameters of the 4WIS experimental platform are summarized in \cref{tab1}.

\subsubsection{Scenario Settings}

The experiments include four representative scenarios: a maze-like environment with narrow corridors and multiple turns (Env1), a structured parking environment with a constrained terminal pose (Env2), a narrow lateral-transfer environment with limited longitudinal clearance (Env3), and an irregular dense-obstacle environment (Env4). These scenarios evaluate global detouring and mode switching, terminal maneuverability and accuracy, lateral-motion capability, and trajectory generation in cluttered spaces, respectively. Together, they provide a comprehensive assessment of the algorithm's overall performance.

For a fair comparison, all algorithms were evaluated using identical robot models and experimental settings. Each scenario was repeated 50 times, and the resulting metrics were statistically analyzed to assess planning efficiency and overall performance.

\begin{table}[t]
\centering
\caption{Key Parameters of the 4WIS Robot and Trajectory Planner}
\label{tab1}
\resizebox{\columnwidth}{!}{%
\begin{tabular}{l @{\hspace{2pt}} c @{\hspace{8pt}} l @{\hspace{2pt}}c}
\toprule
\textbf{Parameter} & \textbf{Value} & \textbf{Parameter} & \textbf{Value} \\
\midrule
\multicolumn{4}{l}{\textit{1. Robot Physical and Kinematic Parameters}} \\
\midrule
Length ($l$) & 1.00 m & Width ($w$) & 0.62 m \\
Wheelbase ($L$) & 0.68 m & Track width ($W$) & 0.52 m \\
Max steering angle ($\phi_\text{max}$) & $30^\circ$ &
Max steering rate ($\dot{\phi}_\text{max}$) & $180^\circ$/s \\
Mode switch time ($t_{\text{switch}}$) & 0.4 s &
Max acceleration ($a_\text{max}$) & 5.0 m/s$^2$ \\
Reference velocity ($v_{\text{ref}}$) & 1.0 m/s &
Max velocity ($v_{\text{max}}$) & 1.0 m/s \\
\midrule
\multicolumn{4}{l}{\textit{2. Front-end Path Planning Parameters}} \\
\midrule
Reverse penalty ($C_{\text{rev}}$) & 2.0 &
Steer-change penalty ($C_{\Delta\text{steer}}$) & 1.0 \\
Steer penalty ($C_{\text{steer}}$) & 1.0 &
Dir-change penalty ($C_{\text{dir}}$) & 1.0 \\
Reference velocity ($v_{\text{ref}}$) & 1.0 m/s &
Sampling time ($\Delta t$) & 0.5 s \\
\midrule
\multicolumn{4}{l}{\textit{3. Back-end Trajectory Optimization Parameters}} \\
\midrule
Resampling distance ($\mathcal{D}$) & 0.2 m & Corridor max distance ($d_\text{max}$) & 5 m \\
Rapid expansion step ($\Delta d_1$) & 0.5 m & Detailed expansion step ($\Delta d_2$) & 0.05 m \\
Comfort cost coefficient ($\alpha$) & 0.1 & Constraint coefficient ($\alpha_{\text{penalty}}$) & $10^\text{6}$ \\
Convergence threshold ($\varepsilon_{\text{tol}}$) & $10^{\text{-6}}$ &
Max iteration times ($\mathcal{I}$) & 10 \\
\bottomrule
\end{tabular}}
\vspace{-5pt}
\end{table}

\subsubsection{Evaluation Metrics}

To comprehensively evaluate the proposed method and the baseline planners, the following metrics are adopted:
\begin{itemize}
    \item \textbf{Safety (SF)}: the collision-free feasibility of the generated trajectory. A trajectory is considered safe only if no collision is detected along its execution.
    \item \textbf{Trajectory Length (TL)}: the total length of the trajectory, computed by accumulating the Euclidean distances between adjacent sampled points.
    \item \textbf{Arrival Time (AT)}: the total duration required by the trajectory to reach the terminal state. 
    \item \textbf{Average Curvature (AC) and Maximum Curvature (MC)}: the average and maximum absolute curvature of the trajectory. The curvature at each intermediate point is estimated from consecutive points:
    \begin{equation}
    \!\!\!\!\!\! \kappa_i
    =
    \frac{2\left| (\boldsymbol{\rho}_i-\boldsymbol{\rho}_{i-1})\times(\boldsymbol{\rho}_{i+1}-\boldsymbol{\rho}_i)\right|}
    {\|\boldsymbol{\rho}_i-\boldsymbol{\rho}_{i-1}\|_2\,\|\boldsymbol{\rho}_{i+1}-\boldsymbol{\rho}_i\|_2\,\|\boldsymbol{\rho}_{i+1}-\boldsymbol{\rho}_{i-1}\|_2},
    \end{equation}
    where \(\boldsymbol{\rho}_i=[x_i,y_i]^{\mathsf{T}}\) is the \(i\)-th sampled position and \(\times\) denotes the scalar cross product.
    \item \textbf{Terminal Position Error (TPE)}: the Euclidean distance between the final trajectory position and the desired goal position.
    \item \textbf{Terminal Yaw Error (TYE)}: the absolute wrapped heading error between the final trajectory heading and the desired goal heading.
    \item \textbf{Standard-mode Rate (SMR)}: the proportion of resampled trajectory samples consistent with a predefined 4WIS motion mode: parallel, Ackermann, lateral Ackermann, in-place rotation, or stop. For each resampled point, let \([v_{x,i}^{b},v_{y,i}^{b},\omega_i]^{\mathsf T}\) denote the body-frame twist. A sample is classified as a standard 4WIS motion if it is approximately stationary or sufficiently close to the parallel, Ackermann, lateral Ackermann, or in-place rotation template. The translational speed, characteristic vehicle radius are defined as: 
    \begin{equation}
    v_i=\sqrt{(v_{x,i}^{b})^2+(v_{y,i}^{b})^2},\ \
    r_v=0.5\sqrt{L^2+W^2}.
    \end{equation}
    With the normalization factor \(q_i=\max\left(v_i,|\omega_i|r_v\right)\), the minimum normalized deviation is defined as:
    \begin{equation}
    e_i^{\min}=
    \min\left(
    \frac{|\omega_i|r_v}{q_i},
    \frac{|v_{y,i}^{b}|}{q_i},
    \frac{|v_{x,i}^{b}|}{q_i},
    \frac{v_i}{q_i}
    \right),
    \end{equation}
    and the classification result is:
    \begin{equation}
    b_i=\mathbb{I}\left[
    \left(v_i<\epsilon_v\land |\omega_i|r_v<\epsilon_v\right)
    \lor e_i^{\min}<\eta
    \right].
    \end{equation}
    where \(\eta=0.15\). A lower SMR indicates more non-standard motions and greater control difficulty.
    \item \textbf{Computation Time (CT)}: the wall-clock runtime of the corresponding planning pipeline.
\end{itemize}

\begin{figure*}[!t]
\centering
\includegraphics[width=0.88\textwidth]{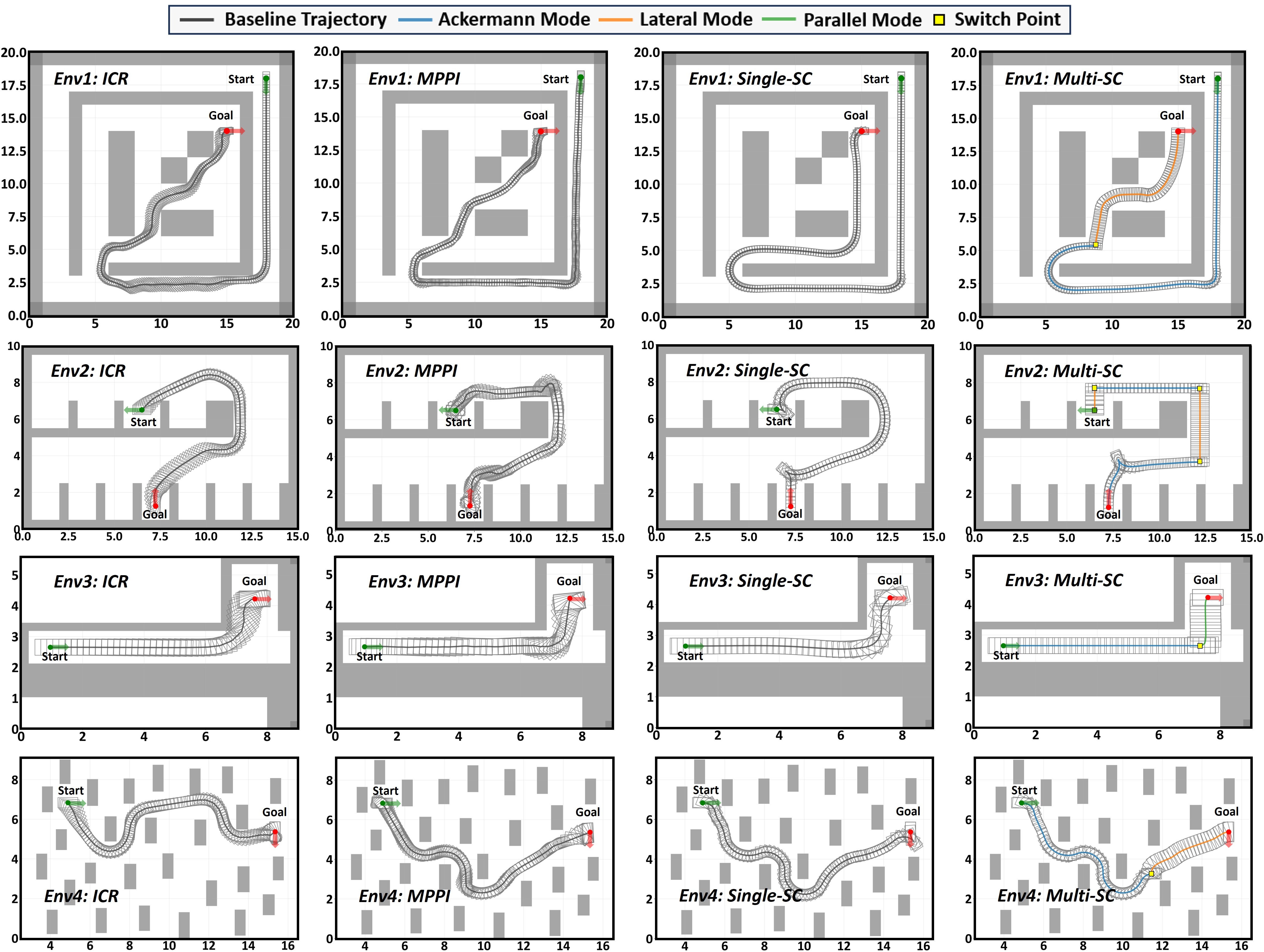}
\caption{Visualization of planned trajectories across the four benchmark scenarios.}
\label{fig5}
\vspace{-5pt}
\end{figure*}

\begin{table*}[!t]
\centering
\caption{Performance comparison of trajectory planning methods across benchmark scenarios.}
\label{tab:benchmark_results}
\renewcommand{\arraystretch}{1.1}
\resizebox{0.88\textwidth}{!}{
\begin{tabular}{llccccccccc}
\toprule
\multirow{2}{*}{\textbf{Scenario}} &
\multirow{2}{*}{\textbf{Method}} &
\multicolumn{9}{c}{\textbf{Experimental Metrics}} \\
\cline{3-11}
& & \textbf{SF} & \textbf{TL (m)} & \textbf{AT (s)}
& \textbf{AC (m$^{-1}$)} & \textbf{MC (m$^{-1}$)}
& \textbf{TPE (m)} & \textbf{TYE (rad)}
& \textbf{SMR} & \textbf{CT (ms)} \\
\midrule
\multirow{4}{*}{Env1}
& ICR & \xmark & 43.623 & 55.008 & 0.256 & \textbf{1.511} & \textbf{0.000} & \textbf{0.000} & 0.728 & 31293 \\
& MPPI & \textbf{\cmark} & \textbf{43.587} & 58.550 & 0.202 & 2.082 & 0.095 & 0.087 & 0.904 & 26097 \\
& Single-SC & \textbf{\cmark} & 48.396 & 49.592 & \textbf{0.186} & 1.695 & \textbf{0.000} & \textbf{0.000} & \textbf{1.000} & 9074.1 \\
& Multi-SC & \textbf{\cmark} & 45.790 & \textbf{46.387} & 0.193 & 1.695 & \textbf{0.000} & \textbf{0.000} & \textbf{1.000} & \textbf{5526.0} \\
\hline
\multirow{4}{*}{Env2} 
& ICR & \xmark & \textbf{15.146} & 19.759 & 0.310 & \textbf{1.616} & \textbf{0.000} & \textbf{0.000} & 0.715 & 2881.8 \\
& MPPI & \textbf{\cmark} & 15.842 & 26.300 & 0.461 & 3.873 & 0.094 & 0.163 & 0.798 & 11626 \\
& Single-SC & \textbf{\cmark} & 17.697 & 18.898 & 0.471 & 1.698 & \textbf{0.000} & \textbf{0.000} & \textbf{1.000} & 3154.9 \\
& Multi-SC & \textbf{\cmark} & 16.762 & \textbf{18.196} & \textbf{0.176} & 1.695 & \textbf{0.000} & \textbf{0.000} & \textbf{1.000} &  \textbf{2628.0} \\
\hline
\multirow{4}{*}{Env3} & 
ICR & \xmark & \textbf{7.6345} & 9.7314 & 0.114 & \textbf{0.830} & \textbf{0.000} & \textbf{0.000} & 0.764 & 6313.9 \\
& MPPI & \textbf{\cmark} & 7.7428 & 10.900 & 0.413 & 2.863 & 0.094 & 0.095 & 0.720 & 4375.3 \\
& Single-SC & \textbf{\cmark} & 8.0603 & 8.8749 & 0.426 & 1.695 & \textbf{0.000} & \textbf{0.000} & \textbf{1.000} & 1542.6 \\
& Multi-SC & \textbf{\cmark} & 7.8961 & \textbf{8.7113} & \textbf{0.102} & 3.142 & \textbf{0.000} & \textbf{0.000} & \textbf{1.000} & \textbf{1025.7} \\
\hline
\multirow{4}{*}{Env4} & 
ICR & \xmark & \textbf{14.054} & 20.429 & 0.571 & 3.646 & \textbf{0.000} & \textbf{0.000} & 0.731 & 26804 \\
& MPPI & \textbf{\cmark} & 14.141 & 25.700 & \textbf{0.560} & 2.400 & 0.097 & 0.056 & 0.839 & 10826 \\
& Single-SC & \textbf{\cmark} & 14.980 & 19.814 & 0.816 & \textbf{1.698} & \textbf{0.000} & \textbf{0.000} & \textbf{1.000} & 2927.9  \\
& Multi-SC & \textbf{\cmark} & 14.243 & \textbf{16.857} & 0.815 & 2.220 & \textbf{0.000} & \textbf{0.000} & \textbf{1.000} & \textbf{1961.8}\\
\bottomrule
\end{tabular}}
\vspace{-5pt}
\end{table*}

\subsection{Comparison Experiments}

\subsubsection{Compared Methods}

To further evaluate the proposed framework, comparative experiments were conducted in the four representative scenarios. The complete method is denoted as Multi-SC, where the front-end performs mode-augmented global search and the back-end conducts trajectory optimization within mode-consistent safe corridors. A single-mode variant, denoted as Single-SC, was also implemented to assess the contribution of multi-modal planning. It uses the same robot geometry and back-end optimizer as Multi-SC, but restricts the front-end search to Ackermann steering.

Two representative planning methods for 4WIS robots were further included as external baselines. The first is an ICR-based 4WIS trajectory planner \cite{add-2}, denoted as ICR, which searches in a general 4WIS motion space and subsequently smooths the resulting path through back-end optimization. The second is the sampling-space-switching model predictive path integral method \cite{add-3}, denoted as MPPI. MPPI is a local sampling-based planning and control method that tracks a front-end reference path while incorporating obstacle-aware costs.

\subsubsection{Experimental Results Analysis}

The trajectory-planning results for all methods and scenarios are visualized in \cref{fig5}, while the corresponding quantitative results are summarized in \cref{tab:benchmark_results}. Overall, Multi-SC achieves the most balanced performance across the four scenarios. The proposed method generates safe trajectories in all cases and reaches the desired terminal pose with zero position and yaw errors at the reported precision. Its SMR consistently remains at 1.000, indicating that the entire trajectory conforms to the predefined standard motion modes and is therefore readily executable. Moreover, Multi-SC achieves the shortest arrival time and computation time in all four scenarios.

The ICR method produces the shortest geometric trajectories in Env2--Env4 and exhibits relatively low maximum curvature in several scenarios. However, because obstacle constraints are not explicitly considered during its back-end smoothing process, the resulting trajectories are classified as unsafe in all four environments. Its SMR ranges only from 0.715 to 0.764, indicating that a considerable proportion of the generated motions cannot be represented by the predefined standard 4WIS modes.

In contrast, MPPI generates collision-free trajectories in all scenarios and generally produces relatively short geometric paths. Nevertheless, it yields the longest arrival time in every scenario, while its computation time is also substantially higher than those of Multi-SC and Single-SC. Since MPPI essentially performs local sampling and tracking based on a front-end reference path without explicitly enforcing an exact terminal-state constraint, it is the only method that exhibits noticeable terminal position and yaw errors. Its SMR ranges from 0.720 to 0.904, suggesting that some trajectory segments contain non-standard coupled motions, which may increase the difficulty of coordinating wheel steering and driving commands.

\begin{figure}[t]
\centering
\includegraphics[width=\columnwidth]{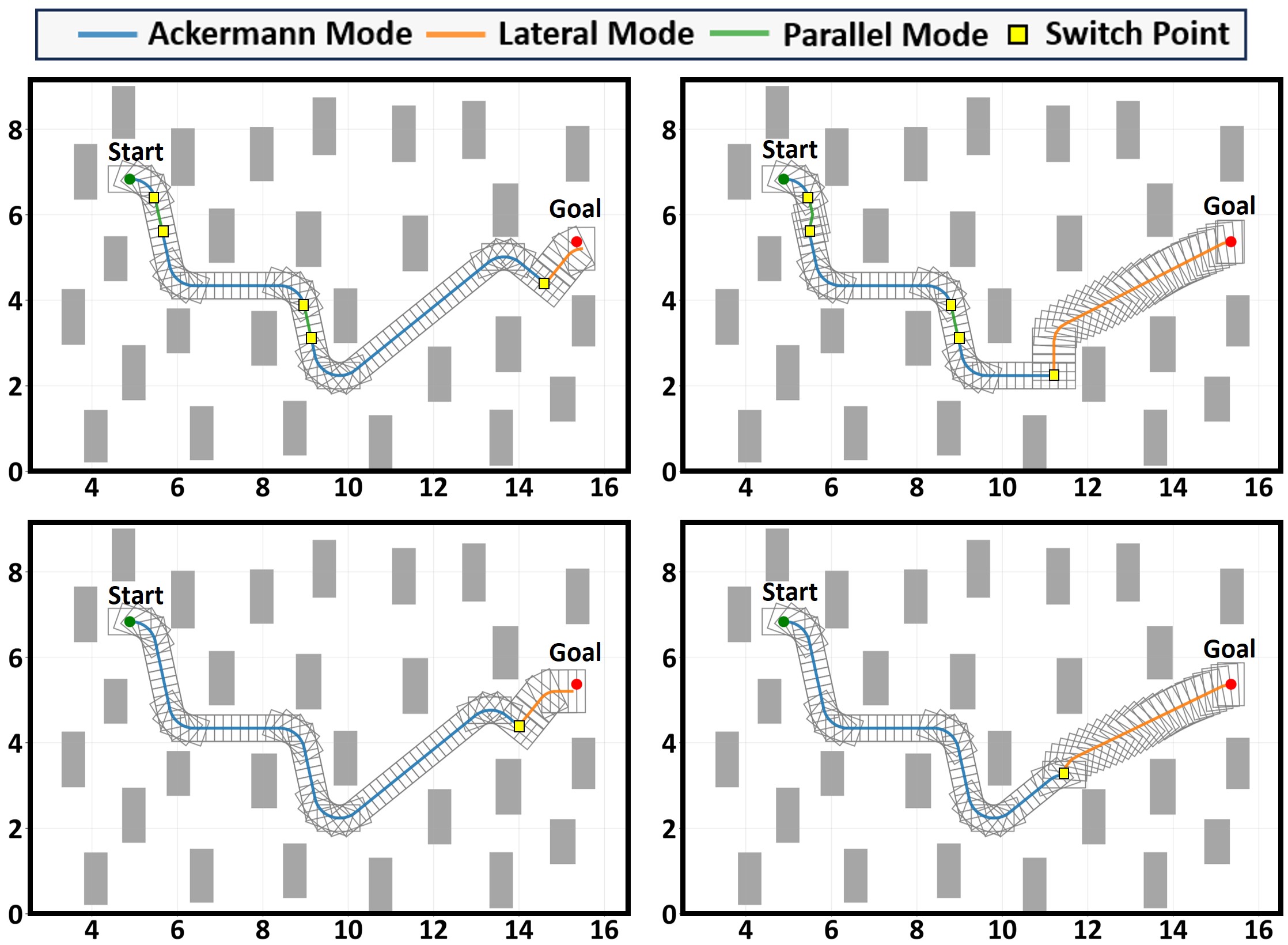}
\caption{Ablation visualization in representative scenario.}
\label{fig7}
\vspace{-5pt}
\end{figure}

Compared with Single-SC, Multi-SC reduces the trajectory length, arrival time, and computation time by an average of 4.4\%, 6.7\%, and 30.6\%, respectively. The multi-modal global search selects lateral or parallel motion according to the environmental structure, thereby avoiding unnecessary detours caused by the restriction to Ackermann steering and reducing both trajectory length and arrival time. Meanwhile, the mode-dependent terminal connection improves front-end search efficiency, while the segment-wise back-end optimization decomposes the complete trajectory into fixed-mode subproblems, thereby reducing optimization complexity. Consequently, despite introducing additional mode-selection decisions, Multi-SC requires less computation time than its single-mode variant and substantially less than the two external baselines.

These results demonstrate that explicitly incorporating multiple steering modes into global planning allows the maneuverability of the 4WIS platform to be more fully exploited. Although each mode transition requires the robot to stop and reconfigure its wheels, the standardized and mode-consistent motion constraints enable stable execution at relatively high speeds during most of the trajectory. Therefore, even when the geometric trajectory generated by Multi-SC is not the shortest, it still achieves a shorter arrival time than the other methods. Overall, the proposed framework provides a better balance among trajectory length, arrival time, obstacle-avoidance safety, terminal accuracy, mode consistency, and computational efficiency.

\subsection{Systematic Analysis}

\subsubsection{Ablation Study of the Front-End Planner}

\begin{table}[t]
\centering
\caption{Ablation comparison in representative scenario.}
\label{tab:ablation_main1}
\renewcommand{\arraystretch}{1.1}
\resizebox{\columnwidth}{!}{
\begin{tabular}{lccccc}
\toprule
\multirow{2}{*}{\textbf{Method}} & \multicolumn{5}{c}{\textbf{Experimental Metrics}} \\
\cline{2-6}
& \textbf{SF} & \textbf{TL (m)} & \textbf{TPE (m)} & \textbf{TYE (rad)} & \textbf{CT (ms)} \\
\midrule
Only MP & \cmark & 18.700 & 0.202 & 0.003 & 116.8 \\
MP+RS & \cmark & 18.745 & \textbf{0.000} & \textbf{0.000} & 185.0 \\
MP+SCH & \cmark & 15.500 & 0.195 & 0.003 & 131.8 \\
Full & \textbf{\cmark} & \textbf{15.235} & \textbf{0.000} & \textbf{0.000} & \textbf{35.44} \\
\bottomrule
\end{tabular}
}\vspace{-5pt}
\end{table}

To verify the contribution of the main front-end modules, an ablation study is conducted in Env4, which has the most cluttered obstacle distribution among the selected benchmark scenarios. Four variants are compared: \textbf{Only MP}, which retains only the multi-modal motion primitives; \textbf{MP+RS}, which further introduces the multi-modal RS heuristic and terminal connection but disables the mode-switching heuristic and switching cost; \textbf{MP+SCH}, which keeps the mode-switching heuristic and switching cost but disables the terminal connection; and \textbf{Full}, which enables all modules.

The results are shown in \cref{fig7} and summarized in \cref{tab:ablation_main1}. The equivalent cost of mode switching is included in the reported trajectory length. Only MP already provides the planner with Ackermann, lateral Ackermann, and parallel-motion reachability, but it lacks both terminal connection and switching regulation, leading to a terminal position error of \(0.202\,\mathrm{m}\) and the largest equivalent trajectory length. MP+RS eliminates the terminal error by introducing the RS-based terminal connection, but without switching regulation it still produces frequent mode transitions, resulting in an equivalent trajectory length of \(18.745\,\mathrm{m}\) and the highest computation time. MP+SCH reduces the equivalent trajectory length to \(15.500\,\mathrm{m}\) by suppressing unnecessary switching, but the terminal position error remains \(0.195\,\mathrm{m}\) because the terminal connection is disabled.

In comparison, the full method achieves zero terminal position and yaw errors, the shortest equivalent trajectory length and the lowest computation time. These results indicate that the multi-modal motion primitives provide basic reachability, the RS terminal connection improves terminal accuracy, and the mode-switching heuristic and cost improve mode-selection efficiency. Combining these modules yields a more compact, accurate, and computationally efficient front-end planning result.

\begin{figure*}[t]
\centering
\includegraphics[width=0.8\textwidth]{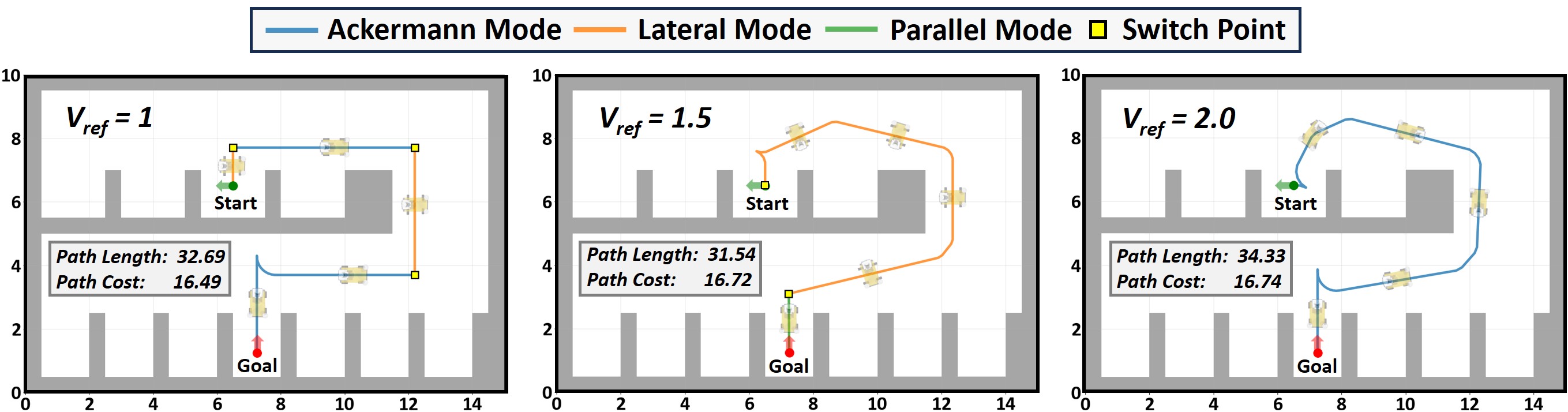}
\caption{Effect of reference velocity on motion mode switching behavior.}
\label{fig6}
\vspace{-5pt}
\end{figure*}

\subsubsection{Sensitivity to the Reference Velocity}

Since the reference velocity \(v_{\mathrm{ref}}\) affects the relative mode-switching cost in the front-end search, different values may lead to different motion-mode preferences. To illustrate this effect, the planner was evaluated in the same structured scenario using three representative reference velocities.

As shown in \cref{fig6}, when \(v_{\mathrm{ref}}\) = 1.0 m/s, the relatively low switching penalty encourages the planner to flexibly combine Ackermann, lateral, and parallel motions. At \(v_{\mathrm{ref}}\) = 1.5 m/s, unnecessary mode transitions are suppressed while the advantage of lateral motion is retained. When \(v_{\mathrm{ref}}\) = 2.0 m/s, mode selection becomes more conservative, and the resulting path is dominated by Ackermann motion.

These results demonstrate that \(v_{\mathrm{ref}}\) provides an interpretable trade-off between multi-modal flexibility and mode-transition conservativeness. Smaller values encourage more frequent use of different motion modes, whereas larger values favor trajectories with fewer mode transitions and stronger mode consistency.

\subsection{Results of Real-world Experiments}

\begin{figure*}[t]
\centering
\includegraphics[width=0.9\textwidth]{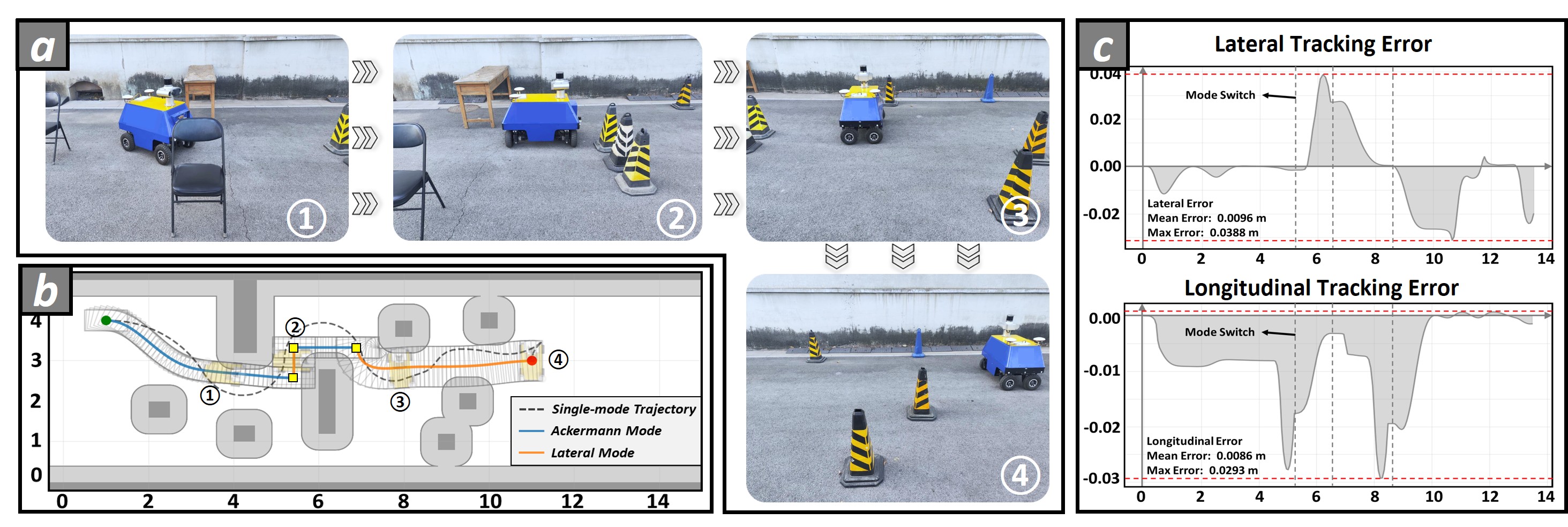}
\caption{Real-world experiment results. (a) Experimental environment and navigation process. (b) Comparison between the single-modal and multi-modal trajectories. (c) Lateral and longitudinal tracking errors during execution.}
\label{fig8}
\vspace{-5pt}
\end{figure*}

To validate the sim-to-real transferability of the proposed framework, real-world planning and execution experiments were conducted, as shown in \cref{fig8}(a). Static obstacles were placed in the environment, and a global map was constructed using LIO-SAM \cite{liosam}. The platform was required to plan and track a collision-free trajectory toward the target. The platform parameters followed \cref{tab1}.

\cref{fig8}(b) compares the single-modal and multi-modal trajectories in the real-world experiment. The proposed multi-modal trajectory completed the task in 15.09 s, whereas the single-modal trajectory required 18.46 s, demonstrating improved motion efficiency by exploiting mode adaptability. The front-end planning and back-end optimization times of the proposed method were 0.82 s and 0.57 s, respectively, resulting in a total pre-execution planning time of 1.39 s. Since the proposed framework acts as a global planner, the complete trajectory is generated before execution.

To evaluate trajectory executability, a simple MPC controller constructed from the kinematic model in \cref{motion1} was used for real-time tracking at approximately 20 Hz. As shown in \cref{fig8}(c), after removing the mode-switching intervals, the average and maximum lateral errors were 0.0096 m and 0.0388 m, respectively, while the average and maximum longitudinal errors were 0.0086 m and 0.0293 m. These results indicate that the proposed framework can generate executable trajectories efficiently and support accurate execution on the physical 4WIS platform.

\section{Conclusion}
This paper proposes a multi-modal global trajectory planning framework for 4WIS robots that bridges discrete multi-modal path generation and executable trajectory construction. In the front-end, Hybrid A* is extended to a four-dimensional state space with explicit motion-mode modeling and mode-switch-aware heuristic and cost functions, together with multi-modal RS curves and an intelligent terminal connection strategy. In the back-end, a mode-consistent segment-wise trajectory optimization framework based on an improved iterative safe corridor scheme converts discrete multi-modal paths into smooth, kinematically feasible trajectories with stationary mode transitions. Experimental results demonstrate that the proposed framework generates safe trajectories conforming to the predefined motion modes, achieves zero terminal pose error, and provides the shortest arrival and computation times among the compared planners. Real-world experiments further verify the executability and practical effectiveness of the generated trajectories. Overall, the proposed framework provides an effective solution for exploiting the maneuverability of 4WIS robots in constrained and structured environments.

Future work will focus on two directions. First, a post-processing module will be introduced to optimize the necessity and locations of motion mode switching points after front-end path planning, further improving trajectory quality. Second, the safe corridor construction will be extended from axis-aligned boxes to general convex polytope representations, aiming to reduce conservativeness while preserving feasibility.

\bibliographystyle{IEEEtran}
\bibliography{ref}

\end{document}